\pdfoutput=1
\documentclass{article}

\usepackage{arxiv}

\usepackage[utf8]{inputenc} 
\usepackage[T1]{fontenc}    
\usepackage{hyperref}       
\usepackage{url}            
\usepackage{booktabs}       
\usepackage{amsfonts}       
\usepackage{nicefrac}       
\usepackage{microtype}      
\usepackage{lipsum}		
\usepackage{graphicx}
\usepackage{doi}
\usepackage{tikz}
\usepackage{amsmath}
\usepackage{float}
\usepackage[numbers]{natbib}

\title{Asterisk*: Keep it Simple}

\hypersetup{
    pdfauthor={Andrew Semenov},
    pdftitle={ASTERISK*: Keep it Simple},
    pdfkeywords={Text Embeddings, Knowledge Distillation, Zero-Shot Classification, Transformer Model},
    pdfsubject={Machine Learning, Natural Language Processing, Model Efficiency},
    pdfproducer={LaTeX},
    pdfcreator={pdflatex}
}

\date{}	

\author{ \href{https://orcid.org/0009-0009-7047-5179}{\includegraphics[scale=0.06]{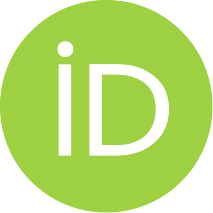}\hspace{1mm}Andrew Semenov} \\
	\texttt{orcid@casd.pro} \\
}

\renewcommand{\headeright}{}
\renewcommand{\undertitle}{}

\begin{document}
\maketitle

\begin{abstract}
This paper describes Asterisk*, a compact GPT-based model for generating text embeddings. The model implements a minimalist architecture with two layers, two attention heads, and 256 embedding dimensions. Through knowledge distillation from larger pretrained models, we examine the trade-offs between model size and performance while reducing computational and memory requirements. The model is evaluated and optimized primarily for classification tasks, where experimental results demonstrate its capability in zero-shot classification across various downstream applications. With additional configuration, the model's performance approaches, and, sometimes outperforms that of larger architectures on specific classification tasks.
\end{abstract}

\section{Introduction}
Text embeddings are a universal tool for various tasks related to language processing, as they provide meaningful representations of text in a standardized format that is easy to work with. However, to capture the full range of meanings in language, models must be trained on large amounts of data and the models themselves must be sufficiently large and complex to effectively learn language patterns. One technique for training new models with reduced computational and data requirements, as well as optimizing existing ones, is Knowledge Distillation (also known as Knowledge Transfer). This approach enables the effective transfer of knowledge from a larger (teacher) model to a smaller (student) model with minimal performance loss. This paper demonstrates that not only model complexity, dataset size but overall approach can be scaled down and simplified, while still achieving almost state-of-the-art performance.

\section{Setup}
\label{sec:headings}

Asterisk* architecture specifically uses a 256-dimensional embedding space, 2 transformer layers, and bidirectional attention with 2 attention heads per layer, having total of \textbf{14,019,584 }parameters. The embedding layer combines token embeddings with positional embeddings, both initialized with a normal distribution. Each token is mapped to a 256-dimensional vector through the embedding layer. \\

The transformer blocks use pre-norm architecture, meaning layer normalization is applied before both the self-attention and feed-forward components. The multi-head attention splits the 256-dimensional embeddings into 2 heads, each handling 128-dimensional chunks. The feed-forward network expands the 256 dimensions to 512 in its hidden layer, using GELU activation. \\

The model uses the GPT2 tokenizer with added special tokens [MASK] and [PAD], totaling at 50259 tokens. \\

Weight initialization uses Xavier uniform for attention and feed-forward layers, while embeddings use the smaller normal initialization for stability. The attention mask handles variable-length sequences by setting padded positions to negative infinity before the softmax operation.

\begin{figure}[H]
    \centering
    \includegraphics[scale=0.7]{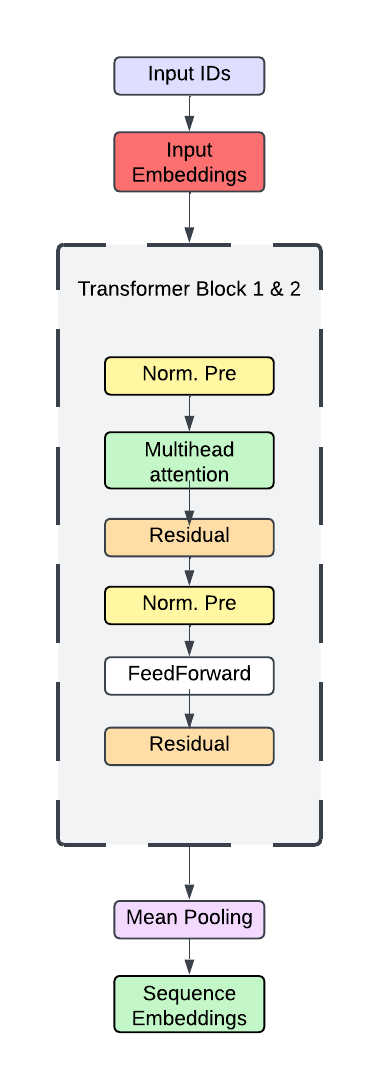}
    \caption{Asterisk* architecture}
    \label{fig:AsteriskArch}
\end{figure}

\subsection{Data}
The model was trained exclusively on English language data. The final iteration of dataset primarily consisted of Wikipedia articles, without any topic-specific selection, and was supplemented with texts from various sources, including academic publications, blogs, fictional books, magazines, TV transcripts, and spoken language. In total, the dataset comprised just \textbf{7 million tokens}, with more than half of these tokens sourced from Wikipedia. All data is publicly available and was collected manually. This relatively small dataset size is compensated by high quality of the contents that cover broad domains of language.

\section{Training}
\subsection{Teacher Model}
The OpenAI text-embedding-3-small model was selected as the teacher model due to its implementation of \href{https://arxiv.org/abs/2205.13147}{Matryoshka Representation Learning (MRL)}. This architectural feature enables dimensional reduction of embeddings from 1536 to 256 dimensions to match Asterisk's* dimensionality requirements while preserving the semantic integrity of the representations. The MRL approach eliminates the need for traditional dimensionality reduction techniques such as Principal Component Analysis (PCA). Our empirical investigations demonstrated that conventional reduction methods introduce significant noise into the compressed embeddings, resulting in substantial degradation of model performance.
\subsection{Loss Function}
For this model the combination of Mean Squared Error (MSE) and Cosine Similarity have been used. The MSE measures the direct distance between student and teacher embeddings, while Cosine Similarity measures the angular difference between embeddings.
\[ \text{Cosine Loss} = 1 - \text{Cosine Similarity} \]
\[
\text{Total Loss} = \alpha \cdot \text{MSE} + (1 - \alpha) \cdot \text{Cosine Loss}
\]
The \(\alpha\) parameter controls the balance between MSE and Cosine loss, higher alpha prioritize absolute distances (MSE), lower alpha prioritize directional similarity (Cosine)
\subsection{ Hyperparameters}

\begin{table}[H]
    \centering
    \begin{tabular}{|c|c|} \hline 
         \(\alpha\)& 0.3\\ \hline 
         Batch Size& 24\\ \hline 
 Learning Rate&0.0001\\ \hline
 Max. Input Sequence Length (in tokens)&128\\\hline
 Optimizer&Adam\\\hline
 Dropout rate&0.1\\\hline
    \end{tabular}
    \caption{Training Hyperparameters}
    \label{tab:hyperparams}
\end{table}
\subsection{Process}
The knowledge distillation process was implemented through an iterative training procedure. For each training step, sequences of 128 tokens were sampled from the dataset and processed in parallel through both the teacher model (OpenAI's text-embedding-3-small) and the student model (Asterisk*) to generate their respective embeddings. The dimensional alignment between the teacher's reduced embeddings (256d) and the student's embeddings enabled direct computation of the distillation loss
\begin{figure}[H]
    \centering
    \includegraphics[scale=0.7]{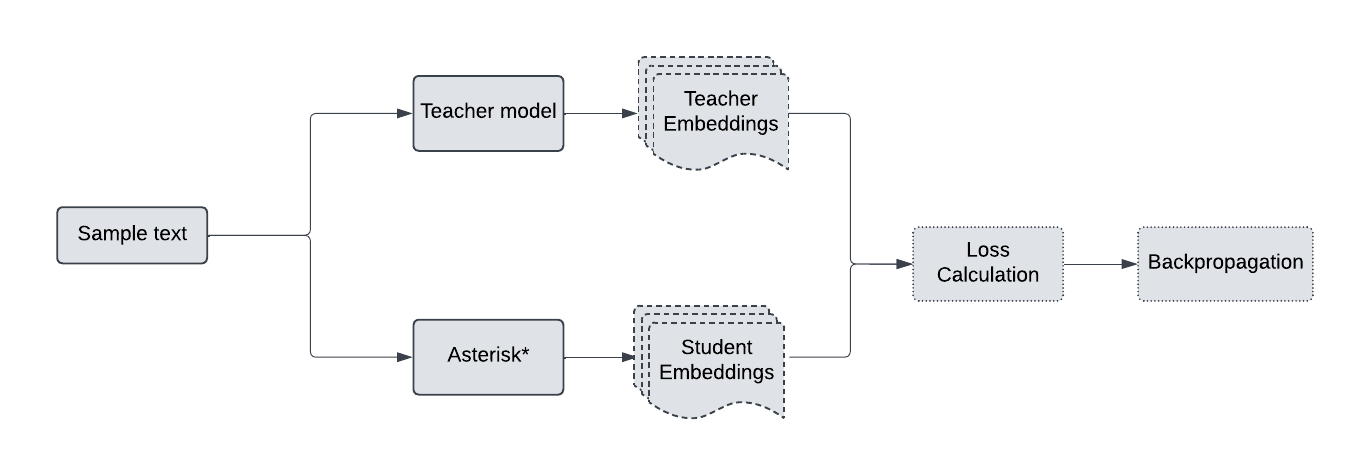}
    \caption{Asterisk* training process}
    \label{fig:Loss}
\end{figure}
The model was trained on a single Nvidia A100 GPU instance, the training took 12 minutes and 41 second for 1 epoch, and \href{https://platform.openai.com/}{OpenAI API} calls being main time consumer.
\begin{figure}[H]
    \centering
    \includegraphics[scale=0.7]{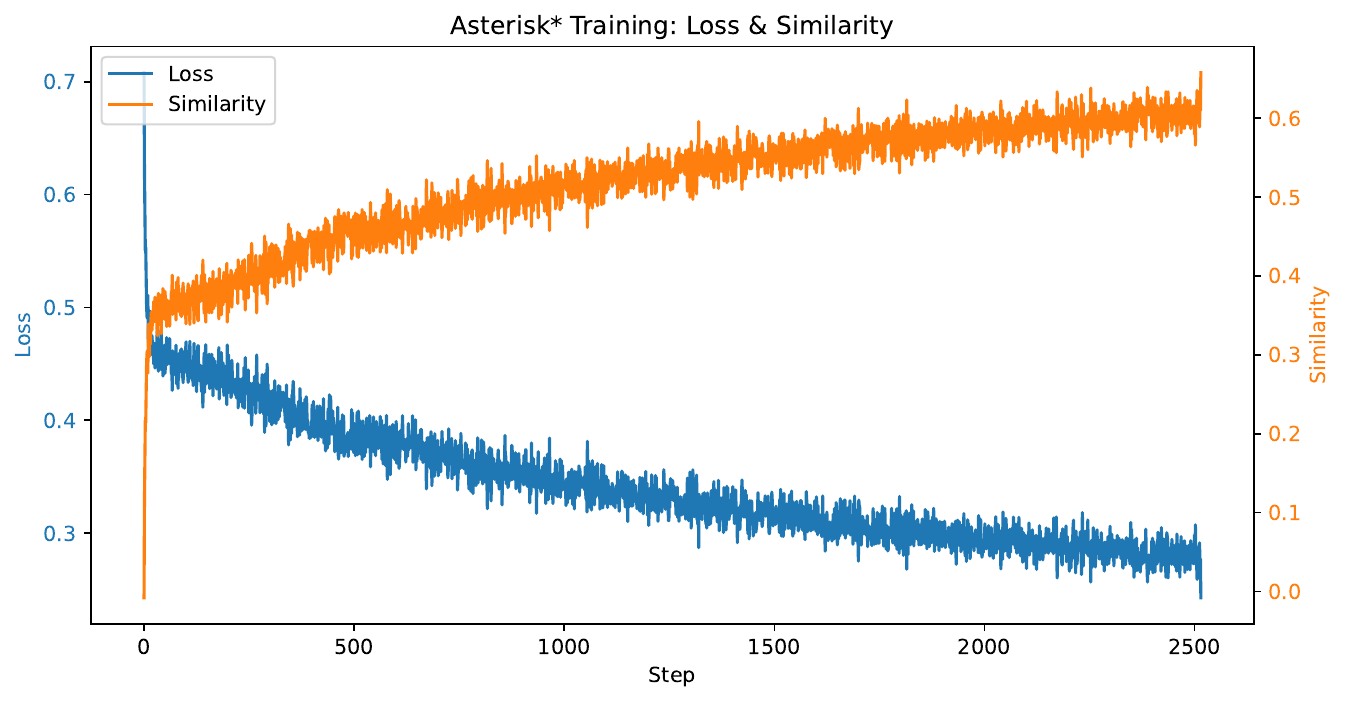}
    \caption{During training, the model achieved cosine similarity with teacher embeddings of 0.65/1 and loss has dropped from 0.7080 to 0.2427}
    \label{fig:AsteriskTrain}
\end{figure}
\section{Evaluations}
During the training process, model checkpoints were preserved at 100-step intervals, resulting in 25 distinct checkpoints. Each checkpoint underwent comprehensive evaluation using both MTEB (Massive Text Embedding Benchmark) benchmarks and custom evaluation methods.\\

Systematic evaluation across all checkpoints revealed that the final checkpoint consistently demonstrated superior performance across all benchmark categories. However, initial analyses highlighted a notable performance disparity in tasks involving informal language processing.\\

Further investigation indicated that this performance bias was primarily attributable to an imbalanced training dataset dominated by formal language samples. Consequently, the model exhibited stronger performance on formal language tasks (e.g., legal document classification) while underperforming on tasks involving natural, informal language (e.g., emotion classification). \\

To address this limitation, was conducted additional fine-tuning using dataset of 5M tokens, comprising Reddit posts and comments, along with conversation transcripts specifically selected for their informal language characteristics. Subsequent evaluation demonstrated improved performance on informal language benchmarks without degrading the model's capabilities on other benchmark categories.
As a baseline for comparison we have used E5-PT Small model from \href{Text Embeddings by Weakly-Supervised Contrastive Pre-training}{Text Embeddings by Weakly-Supervised Contrastive Pre-training} paper.

\subsection{Raw comparison with baseline model}

\begin{table}[H]
    \centering
    \begin{tabular}{|c|c|c|l|}\hline
 & Layers&Hidden Size &Parameters\\\hline \hline 
         Asterisk*&  2& 512 &14M\\ \hline 
         E5-PT Small&  12& 384 &33M\\ \hline
    \end{tabular}
    \caption{Comparison of Asterisk* and baseline model}
    \label{tab:my_label}
\end{table}
\begin{table}[H]
    \centering
\caption{Classification task scores winner in \textbf{Bold}}
\label{tab:my_label}
    \begin{tabular}{|c|c|l|} \hline 
         Benchmark& Asterisk* Score&E5-PT Small\\ \hline 
         MassiveIntentClassification& 46\% &\textbf{70.2\%}\\ \hline
 EmotionClassification&24.7\% &\textbf{42.2\%}\\\hline
 CorporateLobbyingLegalBenchClassification&74.8\% &-\\\hline
 Banking77Classification&20.1\%&\textbf{82.1\%}\\\hline
 MassiveScenarioClassification&51.9\% &\textbf{74.6\%}\\\hline
 AmazonReviewsClassification&23.5\% &\textbf{35.0\%}\\\hline
 ImdbClassificaion&55\% &\textbf{67.9\%}\\\hline
 AmazonCounterfactualClassification&60\% &\textbf{71.7\%}\\\hline
    \end{tabular}
\end{table}

\begin{table}[H]
    \centering
    \begin{tabular}{|c|c|l|} \hline 
         Benchmark& Asterisk* Score &E5-PT Small\\ \hline 
         AskUbuntuDupQuestions& 44.5\% &\textbf{57.8\%}\\ \hline 
 QuoraRetrieval&43.1\% &-\\ \hline 
 StackOverflowQuestions&39.3\% &\textbf{44.4\%}\\ \hline
    \end{tabular}
    \caption{Re-ranking and retrivieval tasks scores}
    \label{tab:my_label}
\end{table}

\subsection{Issues}
During internal evaluation of re-ranking and classification tasks, we observed evidence of probability distribution collapse in the similarity scores between query and candidate embeddings. Specifically, the model demonstrated a tendency to assign elevated similarity scores across text samples, even in cases of limited semantic relevance. However, subsequent detailed analysis revealed that this phenomenon had minimal impact on the model's practical performance metrics.\\

While the model consistently produced higher absolute similarity scores for all candidates during re-ranking and classification tasks, the relative ordering of scores remained preserved, with more semantically relevant texts maintaining higher comparative scores.

\section{Applied Use}

While utilizing raw embeddings for similarity-based classification tasks is a valid approach, it places significant demands on the model's inherent capability to minimize errors. Our internal evaluations revealed that the Asterisk* model, while capable of zero-shot classification using raw embeddings, struggled with complex topics on which it was not sufficiently trained, lacking reliability across a broader range of topics.\\

To address this limitation, we use a Fully-Connected (FC) network architecture on top of the Asterisk* model. This approach was empirically validated to achieve highly reliable classification performance, albeit at the cost of requiring additional task-specific training, unlike the zero-shot setup. Typically the peak performance was achieved after only 1000 samples.\\

The FC network architecture consists of 256 input neurons, with each embedding component assigned to a separate neuron. This is followed by two hidden layers of 128 and 64 neurons, respectively, and an output layer with a neuron count matching the number of target classes. During training we used learning rate of 0.001, batch size of 32  and Cross Entropy Loss (Asterisk* model was not additionally fine-tuned, only FC network was trained during this approach)

\begin{figure}[H]
    \centering
    \includegraphics[scale=0.6]{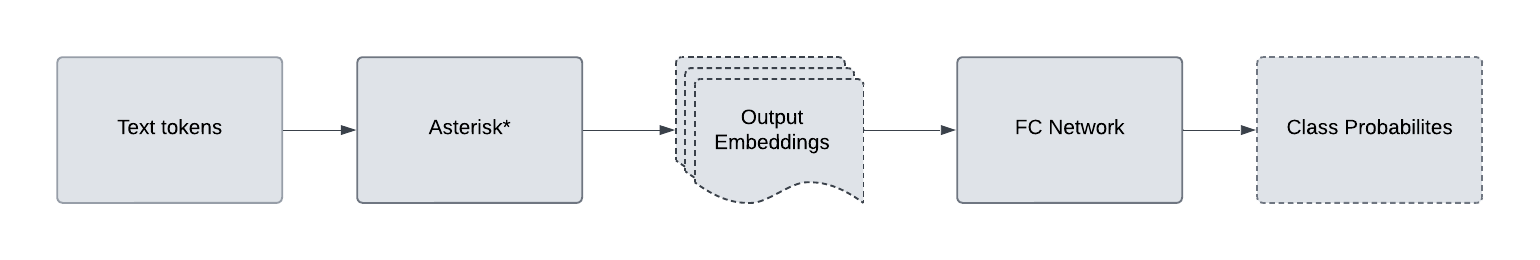}
    \caption{Asterisk* + FC classification setup}
    \label{fig:Loss}
\end{figure}

\subsubsection{MTEB Re-evaluation with FC network}

\begin{table}[H]
    \centering
\caption{Asterisk* + FC, winner in \textbf{Bold}}
\label{tab:my_label}
    \begin{tabular}{|c|c|l|} \hline 
         Benchmark& Asterisk* + FC&E5-PT Small\\ \hline 
         MassiveIntentClassification& \textbf{94\%}&70.2\%\\ \hline
 EmotionClassification&\textbf{76.7\%}&42.2\%\\\hline
 CorporateLobbyingLegalBenchClassification&98.1\%&-\\\hline
 Banking77Classification&75.7\%&\textbf{82.1\%}\\\hline
 MassiveScenarioClassification&\textbf{88\%}&74.6\%\\\hline
 AmazonReviewsClassification&\textbf{62.9\%}&35.0\%\\\hline
 ImdbClassificaion&\textbf{85.6\%}&67.9\%\\\hline
 AmazonCounterfactualClassification&\textbf{81\%}&71.7\%\\\hline
    \end{tabular}
\end{table}
Such setup not only outperforms baseline model across majority of benchmarks but, according to the \href{https://huggingface.co/spaces/mteb/leaderboard}{MTEB Leaderboard}, the Asterisk* model with addition of FC network achieved the following benchmark rankings:

\begin{itemize}
    \item \textbf{1st place} on the \textit{MassiveIntentClassification} task, outperforming 2nd placed model with 7.8 billion parameters
    \item \textbf{2nd place} on the \textit{AmazonReviewsClassification} task, outperforming 3rd placed model with 7.6 billion parameters
\end{itemize}

It is important to note that the MTEB Leaderboard measures the performance of raw, unmodified models, without any additional task-specific components or fine-tuning.

In addition to MTEB re-evaluations we internally simulated possible real world use cases for this model.

\begin{table}[H]
    \centering
    \begin{tabular}{|c|c|l|} \hline 
         Task& Score &Number of classes\\ \hline 
         SMS Spam Classification& 95\% &2\\ \hline
 News Texts Classification& 89\%&5\\\hline
 Sentiment Classification& 87\%&3\\\hline
 Programming Language Classification& 52\%&6\\\hline
 Cancer Doc Classification& 68\%&3\\\hline
 Products Classification& 98\%&4\\\hline
 Offencive Content Detection& 96\%&2\\\hline
    \end{tabular}
    \caption{Performance on internal evaluations using FC network}
    \label{tab:my_label}
\end{table}
The performance of the model using the Fully-Connected (FC) network architecture significantly surpassed what was achievable with raw embeddings alone. This observed improvement in performance can be attributed to the inherent limitations of the Asterisk* model's ability to independently organize conceptual representations within its embedding space.\\

The available training data was likely sufficient for the model to effectively learn the structures and semantic meanings of individual sentences, but insufficient for it to autonomously develop a cohesive, task-agnostic organization of the underlying concepts (at least in most cases). The FC network, appears to provide the necessary capability to interpret the model's embeddings for the target classification tasks.\\

This suggests that while the Asterisk* model's raw embeddings may not be optimally structured for direct similarity-based classification, the information contained within these embeddings can still be effectively exploited through the introduction of a task-specific, learned interpretation mechanism, as implemented by the FC network.

\section{Conclusion}
The methodology presented in this research, while deliberately straightforward in its design, demonstrates that architectural simplicity need not constrain model performance. Our empirical results establish that the addition of a computationally lightweight abstraction layer (FC network) significantly enhances the model's capabilities, surpassing its baseline performance across both standardized benchmarks and practical applications, and comparing to other state-of-the-art models. This research challenges the common assumption that model sophistication necessarily correlates with performance improvement, instead thoughtfully implemented simple architectures and training pipelines can achieve remarkable results while maintaining computational efficiency and design simplicity.

\cite{BERT, TINYBERT, MTEB, DISTILLBERT, GPT, Constr}






\bibliographystyle{plain}
\bibliography{asterisk} 
\end{document}